\documentclass[11pt,a4paper]{article}
\usepackage[hyperref]{acl2020}
\usepackage{times}
\usepackage{latexsym}

\usepackage{graphicx}
\graphicspath{ {./images/}}
\usepackage{microtype}
\usepackage{amsmath}
\usepackage{amssymb}
\usepackage{amsfonts}
\usepackage{amstext}
\usepackage{amsthm}
\usepackage{bbm}
\usepackage{arydshln}

\aclfinalcopy 
\def\aclpaperid{7} 

\title{Attentive Tree-structured Network for Monotonicity Reasoning}

\author{Zeming Chen \\
  Computer Science Department\\
  Rose-Hulman Institute of Technology \\ 
  5500 Wabash Ave, Terre Haute, IN, USA\\
  \texttt{chenz16@rose-hulman.edu} \\}

\date{}

\begin{document}
\maketitle

\begin{abstract}

Many state-of-art neural models designed for monotonicity reasoning perform poorly on downward inference. To address this shortcoming, we developed an attentive tree-structured neural network. It consists of a tree-based long-short-term-memory network (Tree-LSTM) with soft attention. It is designed to model the syntactic parse tree information from the sentence pair of a reasoning task. A self-attentive aggregator is used for aligning the representations of the premise and the hypothesis. We present our model and evaluate it using the Monotonicity Entailment Dataset (MED). We show and attempt to explain that our model outperforms existing models on MED.  
\end{abstract}

\section{Introduction}
In this paper, we present and evaluate a tree-structured long-short-term-memory (LSTM) network in which the syntactic information of a sentence is encoded and the alignment between the premise-hypothesis pair is calculated through a self-attention mechanism. Our work builds on the Child-Sum Tree-LSTM from \citet{tai-etal-2015-improved}. We evaluate our model on several datasets to show that it performs well on both upward and downward inference. Particularly, our model demonstrated good performance on downward inference, which is a difficult task for most NLI models.

Natural language inference (NLI), also known as recognizing textual entailment (RTE) is one of the important benchmark tasks for natural language understanding. Many other language tasks can benefit from NLI, such as question answering, text summarization, and machine reading comprehension. The goal of NLI is to determine whether a given premise \textbf{P} semantically entails a given hypothesis  \textbf{H} \citep{series/synthesis/2013Dagan}. Consider the example below:

\begin{itemize}
  \item {\footnotesize  \textbf{P}: An Irishman won the Nobel prize for literature.}
  \item {\footnotesize  \textbf{H}: An Irishman won the Nobel prize.}
\end{itemize}

\noindent The hypothesis can be inferred from the premise and therefore the premise entails the hypothesis. To arrive at a correct determination, an NLI model often needs to perform different inferences including various types of lexical and logical inferences. In this paper, we are concerned with monotonicity reasoning, a type of logical inference that is based on word replacement. Below is an example of monotonicity reasoning:
 \begin{enumerate}
   \item
   \begin{enumerate}
     \item {\footnotesize \textbf{All} \underline{students}$\downarrow$ carry a \underline{MacBook}$\uparrow$.}
     \item {\footnotesize All students carry a \underline{laptop}.}
     \item {\footnotesize All \underline{new students} carry a MacBook.}
   \end{enumerate}
   \item
   \begin{enumerate}
     \item {\footnotesize \textbf{Not All} \underline{new students}$\uparrow$ carry a laptop.}
     \item {\footnotesize Not All \underline{students} carry a laptop.}
   \end{enumerate}
 \end{enumerate}

\noindent An upward entailing phrase ($\uparrow$) can allow inference from (1a) to (1b), where a more general concept laptop replaces the more specific \textit{MacBook}. A downward entailing phrase ($\downarrow$) allows an inference from (1a) to (1c), where a more specific context \textit{new students} replaces the word \textit{students}. The direction of the monotonicity can be reversed by adding a downward entailing phrase like "Not"; thus (2a) entails (2b).

Recently, \citet{yanaka-etal-2019-neural} constructed a new dataset called the Monotonicity Entailment Dataset (MED). The purpose of that dataset is to evaluate the ability of a neural inference model to perform monotonicity reasoning. It is the first dataset ever created for such purpose. While many neural language models have shown state-of-art performance on large annotated NLI dataset such as the Stanford Natural Language Inference (SNLI) dataset \citep{bowman-etal-2015-large, chen-etal-2017-enhanced, parikh-etal-2016-decomposable}, many of these models did not perform well on monotonicity reasoning. In particular, they had low accuracy when performing downward monotonicity inference. Additionally, most of the state-of-art inference models that do well on upward monotonicity inference perform poorly on downward inference \citep{yanaka-etal-2019-neural}. 


\section{Related Work}
Existing work in this area has adopted a recursive tree-structured neural network for natural language inference. \citet{bowman-etal-2015-recursive} proposed a tree-structured neural tensor network (TreeRNTNs) that can learn representations to correctly identify logical relationships such as entailment. 

\citet{zhou-etal-2016-modelling} extended the recursive neural tensor networks to a recursive long-short term memory network, a tree-LSTM, which combines the advantages of both the recursive neural network structure and the sequential recurrent neural network structure. The tree-LSTM can learn memory cells that reflect the historical memories of the descendant cells and thus improved the model's ability to process long-distance interaction over hierarchies, such as the language parse information. 

\citet{parikh-etal-2016-decomposable} proposed a simple decompose attention model for natural language inference. Their model relies on the attention to decompose the problem into sub-problems so that the smaller problems can be solved separately and in parallel. 

\citet{chen-etal-2017-enhanced}, proposed the Enhanced Sequential Inference Model (ESIM) for natural language inference task. It incorporated the sequential LSTM encoder with the syntactic parsing information from the tree-LSTM structure to form a hybrid neural inference mode. They found that incorporating the parsing information can improve the performance of the model. 

A new type of inference model that relies on external knowledge called the knowledge-based inference model (KIM) was introduced by \citet{chen-etal-2018-neural}. They incorporated neural NLI models with external knowledge in co-attention, local inference collection, and inference composition components. The KIM model achieved state-of-art performance on the SNLI and MNLI datasets.

\section{Our Model}
In this section we present an attentive tree structured network (AttentiveTreeNet) with self-attention based aggregation. This model is composed of the following main components: input sentence embedding, attentive tree-LSTM encoder, self-attention aggregator and a multi-layer perceptron (MLP) classifier. Figure 1 shows the  architecture of our model. Given an input sentence pair, consisting of a premise \textbf{P} and a hypothesis \textbf{H}, the objective of the model is to determine whether \textbf{P} entails  \textbf{H}. 
Our model takes in four inputs: the word embeddings of the premise and hypothesis and the dependency parse trees of the premise and hypothesis. The model initializes the embedding of \textbf{P} and \textbf{H} with some pre-trained word embedding; the parse trees are produced by a dependency parser. Our model forms a Siamese neural network structure \citep{siamese-recurrent}, in which the premise and the hypothesis are passed into a pair of identical tree-LSTMs that share the same parameters and weights. The main idea is to find a function that can map the input sentences into a target space such that we can approximate the semantic distance in the input space.

\begin{figure}[t]
    \centering
    \includegraphics[width=7cm]{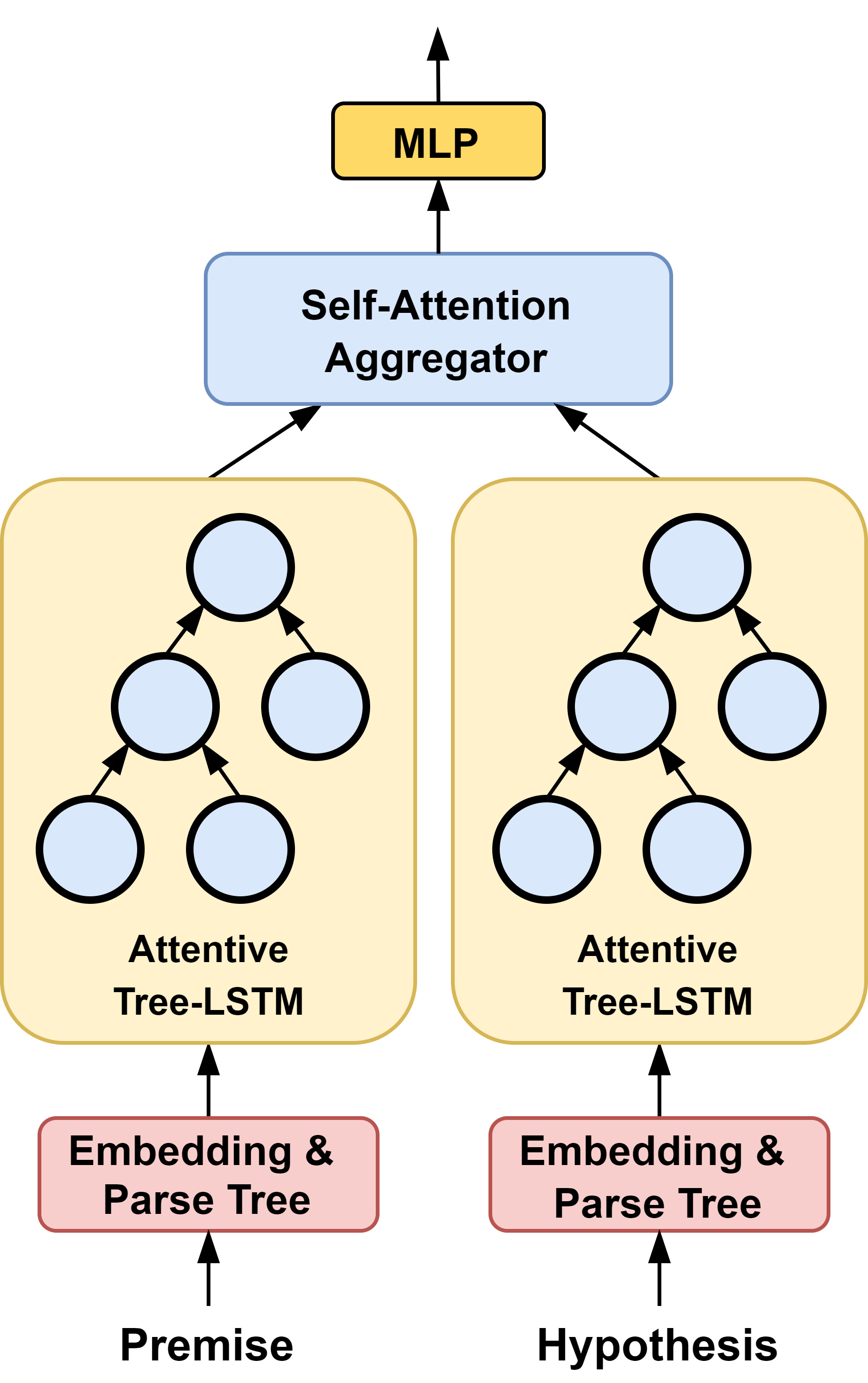}
    \caption{Architecture of our model.}
\end{figure}

\subsection{Attentive Tree-LSTM Encoder}
\paragraph{Child-Sum Tree-LSTM} 
We employ Child-Sum Tree-LSTMs \citep{tai-etal-2015-improved} as the basic building blocks for our model. A standard sequential LSTM network only permits sequential information propagation. However, the \textit{lingistic principle of compositionality} states that an expression's meaning is derived from the meanings of its parts and of the way they are syntactically combined \citep{Partee2007CompositionalityAC}. A tree-structured LSTM network allows each LSTM unit to be able to incorporate information from multiple children units. This takes advantage of the fact that sentences are syntactically formed bottom-up tree-structures. 

A Child-Sum Tree-LSTM is a type of tree-LSTM which contains units that conditioned their components on the sum of their children's hidden states. While a standard sequential LSTM network computes the current hidden state from the current input and the previous hidden state, a child-sum tree-LSTM computes the hidden state from the input and the hidden states of an arbitrary number of children nodes. This property allows relation representations of non-leaf nodes to be recursively computed by composing the relations of the children, which can be viewed as natural logic for neural model \citep{maccartney-manning-2009-extended, zhao-etal-2016-textual}. Using the child-sum tree structure is beneficial in interpreting the entailment relations between parts of the two sentences.

When encoding the sentence in a forward manner, hidden states are passed recursively in a bottom-up fashion. The information flow in each LSTM cell is controlled by a gating mechanism similar to the one in a sequential LSTM cell. The computations in an LSTM cell are as follows:
\begin{align*} \label{eq1}
    \tilde{h} &= \Sigma_{1 \leq k \leq n}h_{k}, \\
    i &= \sigma(W^{(i)}x+U^{(i)}\tilde{h}+b^{(i)}), \\
    o &= \sigma(W^{(o)}x+U^{(o)}\tilde{h}+b^{(o)}), \\
    u &= tanh(W^{(u)}x+U^{(u)}\tilde{h}+b^{(u)}), \\
    f_{k} &= \sigma(W^{(f)}x+U^{(f)}h_{k}+b^{(f)}), \\
    c &= i \odot u+\Sigma_{1 < n}f_{k} \odot c_{k}, \\
    h &= o \odot tanh(c),
\end{align*}

\noindent Here, $k$ is the number of children of the current node, and $\tilde{h}$ is the sum of the hidden states from the children of the current node. The forget gate $f_{k}$ controls the amount of memory being passed from the {\it k}th child. The input gate {\it i} controls the amount of internal input {\it u} being updated and the output gate {\it o} controls the degree of exposure of the memory. The $\sigma$ is the sigmoid activation function, $\odot$ is the element-wise product and {\it W} and {\it U} are both trainable weights to be learned.

\begin{figure}[t]
\centering
\includegraphics[width=7cm]{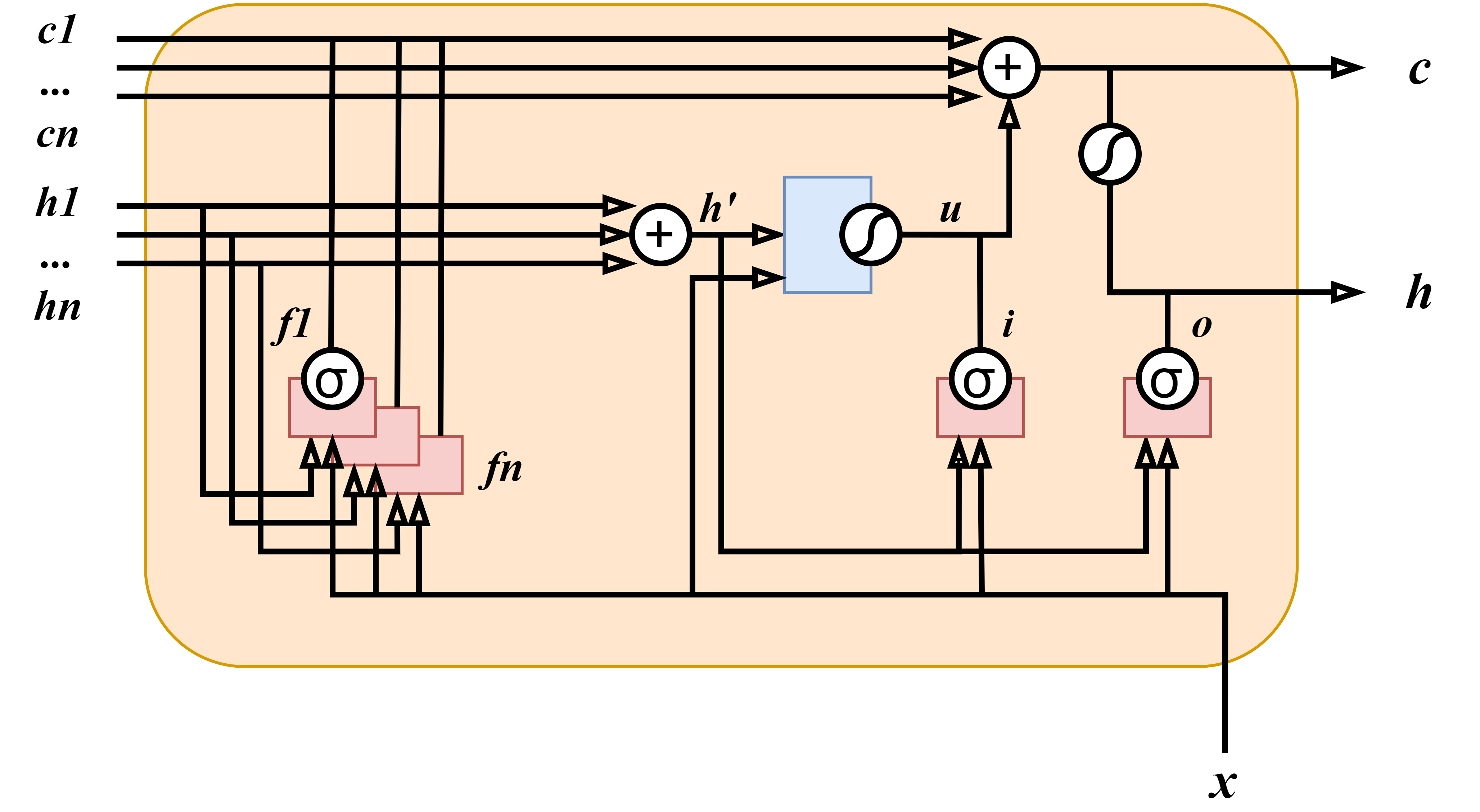}
\includegraphics[width=7cm]{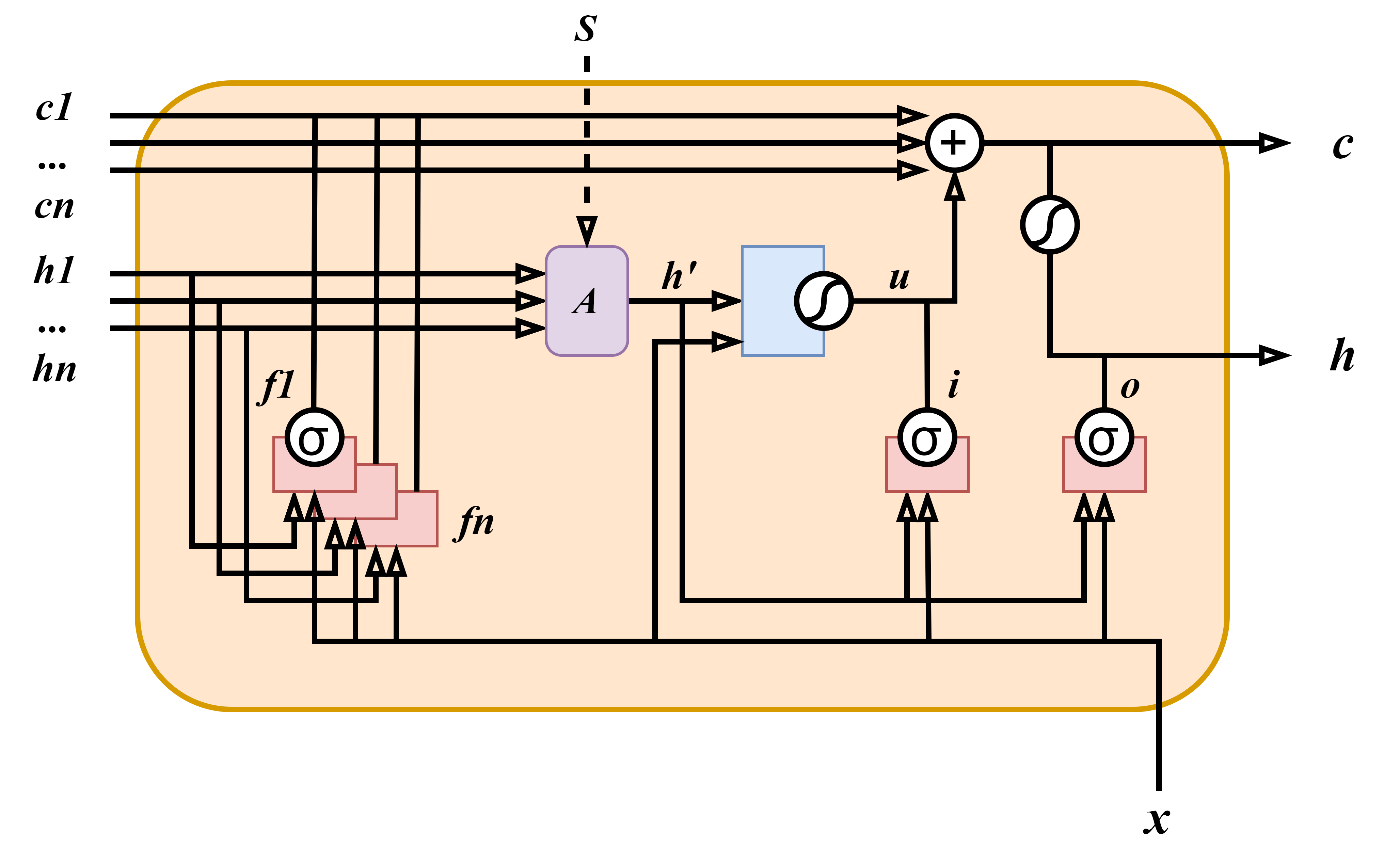}
\caption{A comparison between a standard LSTM cell and an attentive LSTM cell.}
\end{figure}

\paragraph{Attentive Tree-LSTM}
In our model, the standard tree-LSTM is extended to an attentive tree-LSTM \citep{zhou-etal-2016-modelling} by incorporating the attention mechanism into the LSTM cell. In a sentence, some words are more related to the overall context of the sentence than others. The benefit of applying attention is that it considers this semantic relevance by weighting each child according to how relative that child is to the given context. The attention mechanism can assign a higher weight to a child node that is more relevant to the context of the sentence and a lower weight to a child node that is not relevant to the context. 

To apply the attention mechanism, a common soft-attention layer is used in the model. That layer receives a set of hidden states $\{h_{1},h_{2},...,h_{n}\}$ and an external vector {\it s}, which is a vector representation of a sentence from a layer of sequential LSTM. The layer then computes a weight $\alpha$ for each hidden state, and sums up the product of each hidden state and its weight to output the context vector {\it g}. Below are the equations for the soft-attention layer:

\begin{align*}
m_{k} &= tanh(W^{(m)}h_{k} + U^{(m)}s), \\
\alpha_{k} &= \frac{exp(w^{\top} m_{k})}{\Sigma_{j=1}^{n}exp(w^{\top} m_{j})}, \\
g &= \Sigma_{1 \leq k \leq n}\alpha_{k}h_{k}
\end{align*}

\noindent A new previous hidden state is then computed through a transformation $\tilde{h} = tanh(W^{(a)}g + b^{(a)})$. Figure 2 illustrates the standard tree-LSTM cell and the attentive tree-LSTM cell. 

\begin{figure}[t]
\centering
\includegraphics[width=7cm]{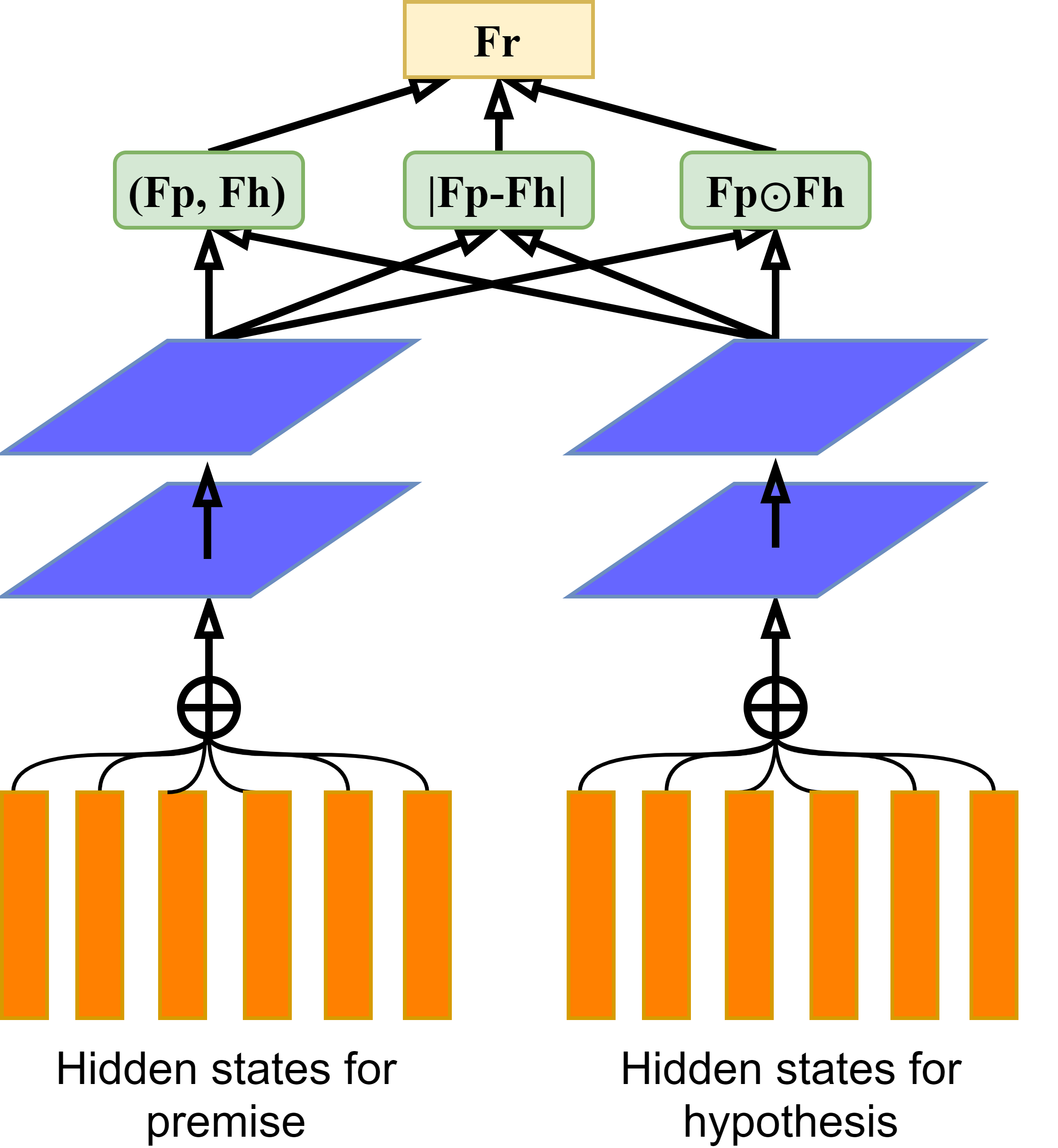}
\caption{Detailed view of the self-attention aggregator}
\end{figure}

\subsection{Self-Attention Aggregator}
After both the premise and the hypothesis are encoded through the tree-LSTM, each tree's hidden states from the nodes are concatenated into a pair of matrices $H_p$ and $H_h$ and passed to a self-attentive aggregator. The aggregator contains a multi-hop self-attention mechanism \citep{Lin2017ASS}.
A sentence has multiple components such as groups of related words and phrases to form an overall context, especially for long sentences. By performing  multiple hops of attention, the model can get multiple attentions that each focus on different parts of the sentence. Given a matrix $H$, the self-attention mechanism performs multiple hops of attention and outputs an annotation matrix $A$ which consists of the weight vector from each hop. $A$ is calculated from a 2-layer multi-layer perceptron (MLP) and a softmax function. Below is the equation to calculate $A$:

\begin{align*}
    A = softmax(W_{s2}tanh(W_{s1}H^{\top}))
\end{align*}
The annotation matrix $A$ is then multiplied by the hidden state matrix $H$ to obtain a context matrix: $M = AH$. In the model, there will be a pair of context matrices $M_p$ and $M_h$. A batch dot product and a {\it tanh} function is then applied to the context matrices with a trainable weight to obtain a pair of output $F_p$ and $F_h$ matrices:
\begin{align*}
    F_p &= tanh(bmm(M_p, W_f)), \\
    F_h &= tanh(bmm(M_h, W_f))
\end{align*}
To aggregate $F_p$ and $F_h$, we follow \citet{Conneau_2017}'s generic NLI training scheme, which includes three matching methods: (i) a concatenation of $F_p$ and $F_h$, (ii) an absolute distance between $F_p$ and $F_h$, and (iii) an element wise product of $F_p$ and $F_h$. Results from the three methods are then concatenated to $F_r$ as the factor of semantic relation between the two sentences which can measure how close the two vector representations of the sentence pair are in the target space. This relatedness information will help the classifier to determine whether the hypothesis is entailed by the premise.

\begin{align*}
    F_r &= [F_p; F_h; |F_p - F_h|; F_p \odot F_h],
\end{align*}

\subsection{MLP}
The factor of relation $F_r$ is fed to a classic three layer MLP classifier. The final prediction is a probability $p_\theta$ representing the degree to which the hypothesis is entailed by the premise. It is calculated by a softmax function, which is a standard activation function used to calculate the probability of the input being in a category for multi-way classification tasks:
\begin{align*}
     Y_1 &= ReLU(W_{f_1}F_r + b_{f_1}), \\   
     Y_2 &= \sigma(W_{f_2}Y_1 + b_{f_2}), \\ 
     y_\theta &= softmax(W_{f_3}Y_2 + b_{f_3}), \\
\end{align*}
\noindent For the classification, the binary cross-entropy loss is used as the objective function:
\begin{align*}
    -\sum_c \mathbbm{1}(X,c)log(p(c|X)),
\end{align*}
where $\mathbbm{1}$ is the binary indicator (0 or 1) whether the label c is the correct class for X.  


\begin{table*}[t]
\centering
\begin{tabular}{llllll}
\hline \textbf{Model} & \textbf{Train Data} & \textbf{Upward} & \textbf{Downward} & \textbf{None} & \textbf{All} \\ \hline
BiMPM \citep{ijcai2017-579} & SNLI & 53.5 & 57.6 & 27.4 & 54.6 \\
ESIM \citep{chen-etal-2017-enhanced} & SNLI & 71.1 & 45.2 & 41.8 & 53.8\\
DeComp \citep{parikh-etal-2016-decomposable}& SNLI & 66.1 & 42.1 & \textbf{64.4} & 51.4\\
KIM \citep{chen-etal-2018-neural} & SNLI & 78.8 & 30.3 & 53.1 & 48.0\\
BERT \citep{devlin-etal-2019-bert}& MNLI & \textbf{82.7} & 22.8 & 52.7 & 44.7 \\
BERT \citep{devlin-etal-2019-bert} & HELP+MNLI & 76.0 & 70.3 & 59.9 & 71.6 \\
AttentiveTreeNet (ours) & MNLI & 54.7 & 60.4 & 37.8 & 58.6 \\
AttentiveTreeNet (ours) & HELP & 55.7 & 72.6 & 57.9 & 66.0 \\
AttentiveTreeNet (ours) & HELP+SubMNLI & 81.4 & \textbf{74.5} & 53.8 & \textbf{75.7}\\
\hline
\end{tabular}
\caption{\label{font-table} Accuracy of our model and other state-of-art NLI models evaluated on MED.}
\end{table*}

\section{Evaluation}
\label{sec:length}
\subsection{Data} 
Six different types of training data are used to train our model.
Initially, we used the HELP dataset \citep{yanaka-etal-2019-help} to train our model. HELP is a dataset for learning entailment with lexical and logical phenomena. It embodies a combination of lexical and logical inferences focusing on monotonicity. HELP consists of 36K sentence pairs including those for upward monotone, downward monotone, non-monotone, conjunction, and disjunction. Next we trained our model with the Multi-Genre NLI Corpus (MNLI) dataset \citep{MultiNLI}. MNLI contains 433k pairs of sentences annotated with textual entailment information. That dataset covers a wide range of genres of spoken and written language. The majority of the training examples in that dataset is upward monotone. In order to provide more balanced training data, we combined a subset of the MNLI dataset with the HELP dataset to reduce the effect of the large number of downward monotone examples in the HELP dataset, we call this combined training data HELP+SubMNLI. The fourth training data contains both the HELP+SubMNLI training data and the training set for simple monotonicity from \citet{richardson2019probing}'s Semantic Fragments. The fifth training data contains both the HELP+SubMNLI training data and the training set for hard monotonicity from Semantic Fragments. Finally, the last training data contains the HELP+SubMNLI training data and the training set for simple and hard monotonicity from Semantic Fragments.

To validate our model's ability for monotonicity reasoning and to evaluate its performance on  upward and downward inference, the Monotonicity Entailment Dataset (MED) was used \citep{yanaka-etal-2019-neural}, which is designed to examine a model's ability of performing monotonicity reasoning. MED contains 5382 premise-hypothesis pairs including 1820 upward inference examples, 3270 downward inference examples, and 292 non-monotone examples. The sentences in MED cover a variety of linguistic phenomena, including lexical knowledge, reverse, conjunction, disjunction, conditional and negative polarity items. We removed sentence pair with the label "contradict" from MNLI dataset since the test dataset MED and the training dataset HELP do not contain the label "contradict". We furthermore tested our model on the simple and hard monotonicity fragments test sets from Semantic Fragments.

\subsection{Training}
Word embeddings are a common way to represent words when training neural networks (Mikolov et al., 2013). To train our model we used Stanford's pre-trained 300-D Glove 840B vectors \citep{pennington-etal-2014-glove} to initialize the word embeddings. The Stanford Dependency Parser \citep{chen-manning-2014-fast} was used to parse each sentence in the dataset. The model is trained with the Adam optimizer \citep{kingma2014adam} which is computationally efficient and helps a model to quickly converge to an optimal result. A standard learning rate for Adam, 0.001, is also used. Dropout with a standard rate of 0.5 is applied to the feed-forward layer in the self-attention aggregator and the classifier to reduce the over-fitting of the model. For the number of hops of the self-attention, we used the default 15 hops. The metric for evaluation is accuracy based. The system is implemented using a common deep learning framework, PyTorch and is trained on a GPU for 20 epochs.  


\section{Results}

\begin{table*}[t]
\centering
\begin{tabular}{lllllll}
\hline \textbf{Test} & \textbf{Model} & \textbf{Training Data} & \textbf{Upward} & \textbf{Downward} & \textbf{None} & \textbf{All} \\ \hline

- & Full Model {\footnotesize w/ vector-concat} & HELP & 55.7 & 72.6 & 57.9 & 66.0 \\
1 & --Self-Attentive Aggregator & HELP & 65.1 & 67.1 & 53.7 & 65.7 \\
2 & --Tree-LSTM  & HELP & 36.6 & 65.5 & 94.8 & 49.5 \\
3 & Full Model {\footnotesize w/ mean-dist} & HELP & 59.3 & 71.2 & 46.2 & 65.9 \\
\hline
- & Full Model {\footnotesize w/ vector-concat} & HELP+SubMNLI & \textbf{81.4} & \textbf{74.5} & 53.8 & \textbf{75.7} \\
1 & --Self-Attentive Aggregator & HELP+SubMNLI & 70.5 & 66.9 & 85.6 & 69.1 \\
2 & --Tree-LSTM   & HELP+SubMNLI & 54.7 & 60.4 & 37.8 & 58.6 \\
3 & Full Model {\footnotesize w/ mean-dist} & HELP+SubMNLI & 68.9 & 73.7 & \textbf{91.0} & 73.0 \\
\hline
\end{tabular}
\caption{\label{font-table} This table shows the accuracy of ablation tests trained on HELP and HELP+SubMNLI and tested on MED. Three ablation test were performed: (i) Remove self-attentive aggregator (--Self-Attentive Aggregator), (ii) Replace tree-LSTM with regular LSTM (--Tree-LSTM) (iii) Use mean distance as a matching method (Full Model {\footnotesize w/ mean-dist}). The final model (Full Model {\footnotesize w/ vector-concat}) uses a concatenation of the sentence vectors as one of the matching methods instead of mean distance.}
\end{table*}

\subsection{Overall Performance} In this section, we evaluated our model's ability of performing monotonicity reasoning. Table 1 shows a comparison of the performance of different models on the Monotonicity Entailment Dataset (MED), including our model. The data for all models except for ours was developed by\citet{yanaka-etal-2019-neural} who developed the MED dataset. Our model achieves an overall accuracy of 75.7\%  which outperforms all other models, even a state-of-art language model like BERT. Table 1 shows the ability of different models on performing upward and downward inference. Our attentive tree model performed better on downward inference than other models with an accuracy of 74.5\% . Our model's performance on upward inference outperforms other models except BERT. However, the upward inference accuracy of our model (81.4) is very close to the accuracy of BERT (82.7). We believe the good performance on upward and downward inference is due to considering parse tree information. Furthermore, the accuracy on upward inference increased significantly when trained with a combination of HELP and MNLI (HELP+SubMNLI) then trained only with HELP; the accuracy increased from 55.7 to 81.4 while the downward accuracy did not change much. Such phenomena suggests that adding MNLI to HELP does reduce the effect of the large number of downward monotone examples in the HELP dataset and thus improve the model's ability on upward inference.   

\subsection{Robustness of Model}
To demonstrate the robustness of our model, we experimented with training the model on various datasets. First, the model was trained on the HELP dataset alone. The overall accuracy was 66.0\%, which outperformed other models from Table 1 except BERT trained with HELP+SubMNLI and our model trained with HELP+SubMNLI. Even on downward inference alone our model outperforms all other models with an accuracy of 72.6\%  except our model trained with HELP+SubMNLI. This result indicates that with a rich set of downward monotone examples, the model can learn to better predict a downward inference problem.

We then trained a model with the MNLI dataset alone. It contains a large amount of upward inference examples and only a rare number of downward inference examples. The result shows that the model generalized to the training data, and had an accuracy of 58.6\%  which is still higher than most models from Table 1. Interestingly, the model's performance on downward inference is still better than its performance on upward inference, even though the training dataset contains a large number of upward monotone examples. This suggests that the model is immune to significant change of training data possibly due to the multiple dropout layer added to the aggregator and the classifier which forces a the model to learn more robust features. As Table 1 show, comparing to BERT trained with MNLI along, our model trained with MNLI along has better performance on downward inference than BERT's performance from \citet{yanaka-etal-2019-neural}.

Finally, we trained our model on a combination of the MNLI dataset and the HELP dataset (HELP+SubMNLI). Because of the large number of upward training examples in MNLI, we suspected that the combination would alleviate the effects of this distortion and as such increase the accuracy for upward inference. We selected 20\%  of the complete MNLI dataset due to the long training period. As the results in Table 1 show, our model still performs well on downward inference with 74.5\%  accuracy, it also showed significant improvements on upward inference with an accuracy of 81.4\% . The overall performance also increased substantially to 75.7\% . Compared to the results of BERT trained with HELP+MNLI from \citet{yanaka-etal-2019-neural}, our model performs better on both upward inference and downward inference, and achieves a higher overall accuracy. The result validates our hypothesis that training on a combination of upward and downward monotone sentences can help the model achieve good performance on both upward and downward monotone, and that the use of AttentiveTreeNet is a good choice.

\subsection{Ablation Test}
To further evaluate which part of the model contributed the most for monotonicity reasoning, we performed several ablation tests on the model. The ablation tests were trained with HELP and HELP+SubMNLI separately and the models were evaluated on the MED dataset. The results are shown in Table 2. We will focus our evaluation on the HELP+SubMNLI data.

For ablation test 1, we removed the self-attentive aggregator and built the feature vector for classification right after the tree-LSTM encoder. As Table 2 (--Self-Attentive Aggregator) shows, performance of the model trained on HELP+SubMNLI shows a significant, 6.6 percentage point drop in overall accuracy, a 10.9 percentage point drop in upward inference accuracy and a 7.6 percentage point drop in downward inference accuracy. The results of this test suggest that the self-attentive aggregator is an important component of the model that cannot be removed.

For ablation test 2, we replaced the tree-LSTM encoder with a standard LSTM encoder. Here, we see an even larger drop in performance. As Table 2 (--Tree-LSTM) shows, performance of the model trained on HELP+SubMNLI shows a large, 17.1 percentage point drop in overall accuracy, a 26.7 percentage point drop in upward inference accuracy and a 14.1 percentage point drop in downward inference accuracy. Based on the results, replacing tree-LSTM with standard LSTM has significant negative impact on the model's monotonicity reasoning performance. Thus, tree-LSTM is a major component of the model that cannot be replaced. 

For ablation test 3, we compared two matching methods for aggregating the two sentence vectors. In our final model (Full Model {\footnotesize w/ vector-concat}), we updated the matching method by following the generic NLI training scheme \citep{Conneau_2017}. In it, we concatenate the two sentence vectors with an absolute distance and an element-wise product as the input vector for the classifier. We compared the performance to our original model (Full Model {\footnotesize w/ mean-dist}) which contains the tree-LSTM encoder, the self-attentive aggregator, and the concatenation of an absolute distance, an element-wise product, and a mean distance as the input vector for the classifier. For this ablation test, the results from Table 2 (Full Model {\footnotesize w/ mean-dist}) are mixed, yet important. While the overall accuracy decreases just slightly, by 2.7 percentage points and the downward inference accuracy only decreases by 0.8 percentage points, the accuracy for upward inference decreases by a significant 12.5 percentage points. We believe that these results justify the use of concatenation of the sentence vector pair. 

Overall, the removal of the Tree-LSTM encoder affected the model's performance most. Thus, we conclude that the Tree-LSTM encoder contributes the most to the model's performance on monotonicity reasoning. 

\begin{table}[t]
\centering
\begin{tabular}{lllll}
\hline \textbf{Training Data} & \textbf{SF} & \textbf{HF} & \textbf{MED} \\ \hline
\multicolumn{4}{c}{{\small Pre-Trained Models}}\\
HELP & 57.0 & 56.8 & 66.0 \\
HELP+SubMNLI & 46.0 & 63.0 & 75.7 \\
\hline
\multicolumn{4}{c}{{\small Re-trained Models w/ SF-training fragments}}\\
HELP{\footnotesize +frag} & 98.1 & 80.6 & 64.5 \\
HELP+SubMNLI{\footnotesize +frag} & 97.8 & 74.8  & 81.5 \\
\hline
\multicolumn{4}{c}{{\small Re-trained Models w/ HF-training fragments}}\\
HELP{\footnotesize +frag} & 74.3 & 95.6 & 68.9 \\
HELP+SubMNLI{\footnotesize +frag} & 73.9 & 93.2 & 73.3 \\
\hline
\multicolumn{4}{c}{{\small Re-trained Models w/ SF and HF-training fragments}} \\
HELP{\footnotesize +frag} & 96.9 & 94.6 & 64.5 \\
HELP+SubMNLI{\footnotesize +frag} & 96.4 & 98.3 & 75.4 \\
\hline
\end{tabular}
\caption{\label{font-table} This table shows the result of the model tested on MED and the simple monotonicity fragments test set (SF) and hard monotonicty fragments test set (HF) from the Semantic Fragments dataset. The table includes three subsections: (i) test accuracy on the three test sets using models pre-trained on HELP and HELP+SubMNLI; (ii) test accuracy on the three test sets using the model re-trained after adding simple monotonicity training set to HELP and HELP+SubMNLI; (iii) test accuracy on the three test sets using the model re-trained after adding hard monotonicity training set to HELP and HELP+SubMNLI; (iv) test accuracy on the three test sets using the model re-trained after adding both simple and hard monotonicity training sets to HELP and HELP+SubMNLI.}
\end{table}

\subsection{Additional Testings}
To check if our pre-trained model can be generalized to other monotonicity dataset, and to see if the model can be easily trained to master the new dataset while retaining its performance on the original benchmark, we conducted some additional testings on the model. We tested our pre-trained models on the Semantic Fragments test dataset which provides a more in-depth test for an NLI model's performance with semantic phenomena, see \citep{richardson2019probing}. Since our model focuses on monotonicity reasoning, we only selected the simple and hard monotonicity fragments for testing. Additionally, since our models are pre-trained on datasets that only contain two labels: "Entailment" and "Neutral", we removed sentence pairs with the third label "contradict" from the test dataset. 

Table 3 shows the results of our testing. While we show the results for both, the HELP and HELP+SubMNLI data sets, we will focus our discussion again on the data obtained with the HELP+SubMNLI data set.

The top portion of Table 3 shows that the model trained on just HELP+SubMNLI performs poorly on the simple and hard monotonicity fragments. This performance is on par with other state-of-art model's, see \citep{richardson2019probing}. 

The first middle portion on Table 3 shows the results of our model's performance when only the {\it simple} training fragments were added to the HELP+SubMNLI training set. As the data shows, the model masters the simple monotonicity reasoning tests, does well on the hard monotonicity reasoning tests and retains its accuracy on the original benchmark MED. 

The second middle portion of Table 3 shows the results of our model's performance when only the {\it hard} training fragments were added to the HELP+SubMNLI training set. In this case, the model masters the hard monotonicity reasoning tests, does well on the simple monotonicity reasoning tests and again retains its accuracy on the original benchmark MED. 

The bottom portion of Table 3 shows the results of our model's performance when both the {\it simple} and {\it hard} training fragments were added to the HELP+SubMNLI training set. As the results show, the model masters both the simple and hard monotonicity reasoning tests while retaining its accuracy on the original benchmark MED.

Overall, the results show that the model trained on the fragments can be generalized to both simple and hard monotonicity reasoning. 

\section{Conclusions}
In this paper, we explained our attentive tree-structured network to perform monotonicity reasoning. Our model combines a tree-structured LSTM network and a self-attention mechanism, which is a potential mechanism for future natural language inference models, to incorporate syntactic structures of the sentence to improve sentence-level monotonicity reasoning. We evaluated our model and showed that it achieves better accuracy on monotonicity reasoning than other inference models. In particular, our model is performing significantly better on downward inference than others. We interpret the results of the experiments as supporting the thesis that using parse trees of a sentence are helpful in inferring the entailment relation.  

Future research on the attentive tree network might extend a tree-LSTM architecture by replacing the LSTM cell with newer language models that have much better performance on various number of natural language processing tasks. One such model is the transformer model. Furthermore, future work might want to investigate how different attention mechanism affect a model's performance.

\section*{Acknowledgments}
We thank Michael Wollowski for reading and giving feedback on drafts and revisions of this paper. We also thank the anonymous reviewers for providing helpful suggestions and feedback.

\bibliography{anthology, attentive}
\bibliographystyle{acl_natbib}

\end{document}